\documentclass[runningheads]{llncs}
\usepackage{graphicx}
\usepackage{amsmath}
\usepackage{amssymb}
\usepackage{booktabs}
\usepackage{enumitem}

\usepackage{lipsum}
\usepackage{xcolor}
\usepackage{cuted}
\usepackage{comment}
\usepackage{pifont}%
\newcommand{\cmark}{\ding{51}}%
\newcommand{\mathleft}{\@fleqntrue\@mathmargin0pt}

\usepackage{hyperref}

\usepackage[nameinlink,capitalize]{cleveref}
\crefname{section}{Sec.}{Secs.}
\Crefname{section}{Section}{Sections}
\Crefname{table}{Table}{Tables}
\crefname{table}{Tab.}{Tabs.}

\usepackage{xspace}
\newcommand*{\eg}{e.g.\@\xspace}

\newcommand{\bftab}{\fontseries{b}\selectfont}
\newcommand{\myparagraph}[1]{\vspace{.2cm} \noindent \textbf{#1:}}
\newcommand{\mysubparagraph}[1]{\vspace{.2cm} \noindent \textit{#1:}}

\begin{document}
\title{DT2I: Dense Text-to-Image Generation from Region Descriptions}

\titlerunning{DT2I: Dense Text-to-Image Generation from Region Descriptions}
\author{Stanislav Frolov\thanks{Equal contribution}\inst{1,2} \and
Prateek Bansal$^{\star}$\inst{1} \and
Jörn Hees\inst{2} \and
Andreas Dengel\inst{1,2}}
\authorrunning{S. Frolov and P. Bansal et al.}

\institute{Technical University of Kaiserslautern, Germany \and German Research Center for Artificial Intelligence, Germany
\email{firstname.lastname@dfki.de}\\}

\maketitle              %
\begin{abstract}
Despite astonishing progress, generating realistic images of complex scenes remains a challenging problem.
Recently, layout-to-image synthesis approaches have attracted much interest by conditioning the generator on a list of bounding boxes and corresponding class labels.
However, previous approaches are very restrictive because the set of labels is fixed a priori.
Meanwhile, text-to-image synthesis methods have substantially improved and provide a flexible way for conditional image generation.
In this work, we introduce dense text-to-image (DT2I) synthesis as a new task to pave the way toward more intuitive image generation.
Furthermore, we propose DTC-GAN, a novel method to generate images from semantically rich region descriptions, and a multi-modal region feature matching loss to encourage semantic image-text matching.
Our results demonstrate the capability of our approach to generate plausible images of complex scenes using region captions.

\end{abstract}

\section{Introduction}
\label{sec:intro}
In the last few years, deep generative image modelling has experienced remarkable progress \cite{Reed2016,Taming}.
Current models can produce realistic results when trained on single-domain datasets such as human faces, birds or flowers, but struggle when trained on complex datasets with multiple objects such as COCO \cite{COCO} and Visual Genome \cite{VisualGenome}.
While learning the natural image distribution via unconditional image synthesis is interesting, controlling the image generation process is important for many practical applications such as image editing, computer-aided design, and visual storytelling.

Strong supervision in the form of segmentation masks often leads to impressive visual results \cite{pavllo2020controlling}, but they are difficult and time-consuming to create from a user's perspective.
Recently, layout-to-image methods \cite{Layout2ImageIJCV,LostGANv2} have attracted much interest by conditioning on a spatial layout of bounding boxes and class labels to allow the user to create a complex scene of multiple objects.
To gain control over the specific appearance of objects, \cite{AttrLayout2Im,frolov2021attrlostgan} further improved the methods by providing additional attributes to the corresponding objects (\eg ``red bus'').
However, the set of class labels and attributes is fixed a priori which strongly limits their expressiveness.
In contrast to labels, text is much more flexible, intuitive and can carry rich semantic information about the object's appearance and relationship to other objects \cite{frolov2021adversarial}.

In this paper, we introduce dense text-to-image (DT2I) synthesis as a new task with the goal to generate realistic images using multiple region descriptions.
See \cref{tab:task_overview} for an overview of tasks and corresponding inputs.
To solve this task, we propose a novel method based on a state-of-the-art adversarial layout-to-image \cite{LostGANv2} model and incorporate best practices from the text-to-image literature (\eg triplet \cite{Reed2016} and DAMSM \cite{AttnGAN} losses), and propose a novel multi-modal region feature matching loss between real and generated image-text pairs.
We create a synthetic dataset to validate the effectiveness of our approach.
Finally, we extensively evaluate our model on a challenging real-world dataset to demonstrate the capability of our method.
Our model outperforms previous methods on several metrics while allowing free-form regions descriptions as input.

\begin{table}[tb]
\setlength{\tabcolsep}{6pt}
\begin{center}
    \resizebox{%
      \ifdim\width>\columnwidth
        \columnwidth
      \else
        \width
      \fi}{!}{%
    \begin{tabular}{l c c c }
    \toprule
    Input         & layout-to-image (L2I) & text-to-image (T2I) & dense text-to-image (DT2I) \\
    \midrule
    spatial layout &  \cmark & (\cmark)  & \cmark \\
    class labels   &  \cmark & (\cmark)  & (\cmark) \\
    free-form text &  - & \cmark  & \cmark \\
    dense captions &  - & -  & \cmark \\
    \bottomrule
    \end{tabular}}
    \end{center}
    \caption{
    Overview of tasks and their corresponding input.
    Brackets indicate optional implicit information.
    L2I methods use a spatial layout of bounding boxes and class labels, while T2I methods use single captions.
    We propose DT2I as a new task with the goal to produce realistic images from multiple localized free-form region descriptions.
    }
\label{tab:task_overview}
\end{table}

\section{Related Work}
\label{sec:related_work}

\myparagraph{Layout-to-Image Synthesis}
The layout-to-image (L2I) task was first studied in \cite{Layout2Im} using a VAE \cite{VAE} by composing object representations into a scene before producing an image.
It was further improved in \cite{Layout2ImageIJCV} with an object-wise attention mechanism to predict a map of object details.
Adversarial approaches \cite{LostGAN,LostGANv2} were able to produce higher-resolution images and provide better control of individual objects by using a reconfigurable layout with separate latent style codes.
Recent developments focused on better instance representations \cite{OCGAN}, context-awareness \cite{he2021context}, and improving the mask prediction of overlapping and nearby objects \cite{li2021image}.
Recently, \cite{AttrLayout2Im,frolov2021attrlostgan} enabled more explicit appearance control of individual objects by conditioning on attributes.
However, the set of attributes is limited and lacks the ability to model complex interactions between objects.
In contrast, our model generates an image from free-form region descriptions.

\myparagraph{Text-to-Image Synthesis}
Generating images from text descriptions (T2I) is a challenging but fascinating problem with remarkable progress in recent years \cite{frolov2021adversarial}.
Compared to labels, they can carry dense semantic information about the appearance of objects and scenes.
Initial approaches \cite{Reed2016} conditioned on a sentence embedding and trained the discriminator to distinguish between real, matching, and generated, non-matching image-text pairs.
Stacked architectures \cite{StackGAN} were further improved by using attention mechanisms \cite{AttnGAN}, contrastive losses \cite{XMCGAN} and transformers \cite{DALLE}.
However, all previous approaches are either applied on single-object datasets or aim to generate an image from one short text description which lacks fine-grained details.

\myparagraph{Combinations of Location \& Text}
In \cite{reed2016learning}, the location of one object can be specified by a bounding box or keypoints, while the appearance is described by a text description.
Generating complex scenes from a single description is challenging, as it requires modelling multiple objects.
To alleviate this problem, \cite{InferGAN,ObjGAN} used a text-to-layout-to-image framework to first predict a layout of bounding boxes which are subsequently refined into shapes before producing the output image. %
In \cite{Hinz2019GeneratingMO,OPGAN}, an object pathway is added to both generator and discriminator which uses bounding boxes and class labels to focus on the individual objects in the scene. %
Sparse semantic masks are used in \cite{pavllo2020controlling,li2020image} to define the position of objects, while an input text can be used to control the style.
In \cite{koh2021text}, the Localized Narratives \cite{pont2020connecting} dataset is used to generate images from mouse traces and paired descriptions.
Different from all previous works, our model takes free-form captions and corresponding locations as input and hence allows rich expression of individual image regions.

\begin{figure}[tb]
  \centering
    \includegraphics[width=\linewidth]{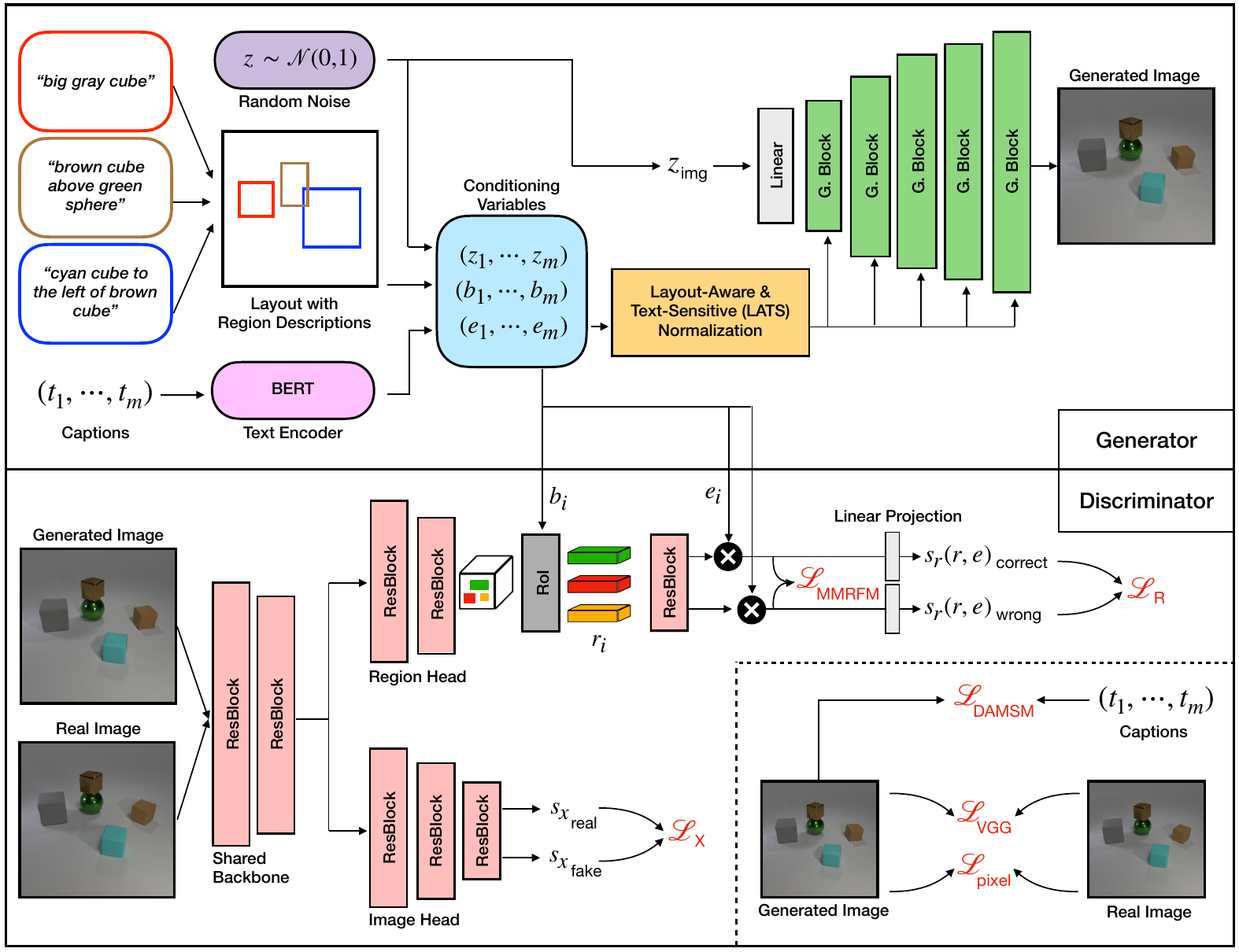}
    \caption{Given region descriptions as input, our novel DTC-GAN is trained to minimize several losses which encourage both image quality as well as image-text alignment.}
    \label{fig:architecture}
\end{figure}

\section{Method}
\label{sec:method}
The goal of our method is to produce realistic images of complex scenes depicting multiple objects from localized region descriptions (which can be seen as the inverse of dense image captioning).
To address this challenging task, we propose a novel framework consisting of the following key components: 1) Dense-Text-Conditional GAN, and 2) Regional Semantic Image-Text Matching.
See \cref{fig:architecture} for an illustration of our DTC-GAN architecture.

\subsection{Dense-Text-Conditional GAN}

\myparagraph{Generator}
Our DTC-GAN builds upon a state-of-the-art L2I model \cite{LostGANv2}.
The generator consists of a linear layer to process the global image latent code $z_\textrm{img} \sim \mathcal{N}(0,1)$ and multiple ResNet \cite{ResNet} based generator blocks with upsampling to produce an image.
To condition our generator on a layout of region descriptions, we adapt the feature normalization technique proposed in \cite{LostGAN}.
Given a layout $L=\{(b_i, t_i)_{i=1}^{m}\}$ of $m$ regions as input, where each region is described by a bounding box $b_i$ and text description $t_i$, we first embed $t_i$ using a pre-trained fixed BERT \cite{Devlin2019BERTPO} model.
Next, we concatenate regional latent codes $Z_\textrm{r}=\{z_i\}_{i=1}^m$ sampled from $\mathcal{N}(0,1)$, and text embeddings $E_\textrm{r}=\{e_i\}_{i=1}^m$ to produce the embedding matrix $\mathcal{S}=(Z_\textrm{r},E_\textrm{r})$ of size $m \times (d_z + d_e)$, with the dimensions $d_z=128$ and $d_e=768$.
As in \cite{LostGANv2}, we use the embedding matrix $\mathcal{S}$ to regress masks for each region using a sub-network.
Finally, we predict affine transformation parameters $\gamma$ and $\beta$ at each generator layer to modulate the visual feature maps in the generator after normalizing them as in BatchNorm \cite{BatchNorm}.
Using the bounding boxes $b_i$, we unsqueeze $\gamma$ and $\beta$ to their corresponding bounding boxes and weight them by the predicted region masks.
Please refer to \cite{LostGANv2} for more details on this normalization technique.
Given that we use text embeddings, our $\gamma$ and $\beta$ parameters are now \textit{Layout-Aware \& Text-Sensitive (LATS)}.

\myparagraph{Discriminator}
The architecture of the discriminator remains largely unchanged from \cite{LostGANv2} and consists of multiple ResBlocks.
A shared backbone takes images as input to extract coarse features, while two classification heads are used for adversarial training to encourage realism on the image and region level.
More specifically, the region head extracts region features using the bounding boxes and RoIAlign operation \cite{he2017mask}, and adopts projection-based conditioning \cite{cGAN,BigGAN} to compute a region-embedding score $s_r$.
The image head continues processing the full image without any semantic knowledge to produce an image score $s_x$.
Both image and region-embedding scores are used for adversarial training.

\subsection{Regional Semantic Image-Text Matching}
The goal of a good DT2I model is not only to produce realistic images but also such that correctly reflect the semantic meaning of the input captions at the corresponding locations.
Inspired by the T2I literature, we adapt two main techniques to learn semantic image-text matching and apply them on the regional level.
Furthermore, we propose a multi-modal region feature matching loss to guide the region classification head of our discriminator.

\myparagraph{Regional Triplet Loss}
The region-embedding classification head in the discriminator takes pairs of extracted region features and corresponding text embeddings as input and is trained to distinguish real from generated pairs.
However, even if the generator produces realistic image regions, there is another kind of possible error, namely, non-matching.
Similar to \cite{Reed2016}, we construct another pair of real image regions and randomly chosen captions to encourage semantic image-text matching and penalize mismatching pairs.

\myparagraph{Regional DAMSM Loss}
The Deep Attentional Multimodal Similarity Model (DAMSM) proposed in \cite{AttnGAN} computes the similarity between an image and global sentence as well as word features using an attention mechanism to improve semantic matching.
We first fine-tune an Inception-v3 \cite{szegedy2016rethinking} image encoder that was pre-trained on ImageNet \cite{ImageNet}, and a pre-trained BERT text encoder to map matching image regions and text features into a common embedding space.
The encoder networks can then be used to provide a fine-grained learning signal to our generator.
We follow the procedure in \cite{AttnGAN}, but in contrast to traditional T2I methods, our captions describe specific regions in the image which leads to better alignment between individual regions and corresponding captions.

\begin{figure}[tb]
  \centering
    \includegraphics[width=0.7\linewidth]{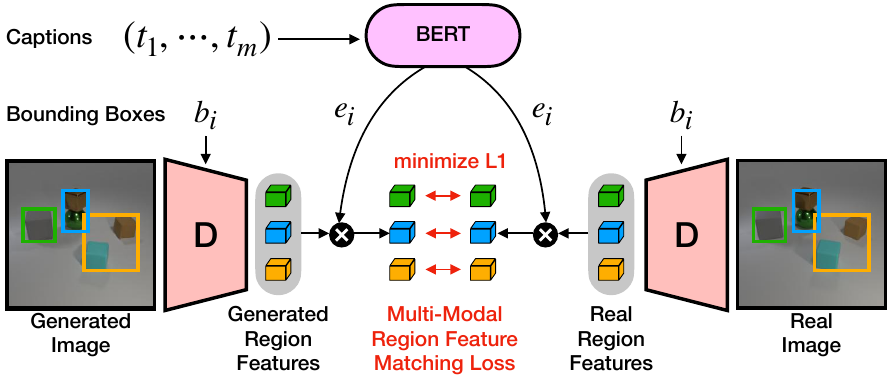}
    \caption{Overview of our Multi-Modal Region Feature Matching loss $\mathcal{L}_{\textrm{MMRFM}}$.
    We first extract region features using bounding boxes $b_i$.
    Next, we multiply individual region features with corresponding text embeddings and minimize L1 distance between real and generated multi-modal region features to improve semantic image-text matching.}
    \label{fig:mmrfm}
\end{figure}

\myparagraph{Multi-Modal Region Feature Matching (MMRFM)}
The regional triplet and DAMSM losses encourage images with matching captions.
However, adversarial losses can lead to adversarial examples and DAMSM lacks proper localization.
Because of the complexity of natural language and diversity of real images, our task is an inherent one-to-many mapping problem between input conditions and output images which pixel-wise losses alone can not handle.
Inspired by the perceptual feature loss \cite{ledig2017photo}, which maximizes similarity at the intermediate feature space, we propose a Multi-Modal Region Feature Matching (MMRFM) loss on the semantics-enriched region features denoted as $\mathcal{L}_{\textrm{MMRFM}}(r_\textrm{real},r_\textrm{fake},t)$.
More precisely, we use the resulting region features after projection-based conditioning and minimize the distance between corresponding real and generated image-text features at the region level, see \cref{fig:mmrfm}.

\subsection{Training Objectives \& Implementation Details}
We combine multiple loss functions during training of our model.
Given an image $x$ and corresponding layout $L$ with text descriptions $t_i$ and regions $r_i$, the discriminator predicts a score for the full image $s_x=D_X(x)$, and matching-aware region-embedding scores $s_r=D_R(r_i, t_i)$.
We consider $(r_i,t_i)$ to be correct if $r_i$ is a real image region with corresponding and matching caption $t_i$, and wrong if a) $r_i$ is generated, or b) $(r_i,t_i)$ is a non-matching pair.
Using an adversarial hinge loss \cite{tran2017hierarchical,BigGAN,LostGAN}, we get $\mathcal{L}_{X}(x)$ as our (unconditional) image loss, $\mathcal{L}_{R}(r, t)$ as our (conditional) regional triplet loss, and the discriminator objective as:

\begin{align}
  \mathcal{L}_{X}&=\mathbb{E}[\max (0,1-D_X(x_\textrm{real})]+\mathbb{E}[\max (0,1+D_X(x_\textrm{fake})] \label{eq:x_hinge}\\
   \begin{split}
    \mathcal{L}_{R}&=\mathbb{E}[\max (0,1-D_R(r_\textrm{real}, t_\textrm{match})] \label{eq:r_hinge}\\
                   &+\mathbb{E}[\max (0,1+D_R(r_\textrm{fake}, t_\textrm{match})] \\
                   &+\mathbb{E}[\max (0,1+D_R(r_\textrm{real}, t_\textrm{non-match})]
   \end{split}\\
  \mathcal{L}_{D}&=\lambda_1\mathcal{L}_{X}+\lambda_2\mathcal{L}_{R} \label{eq:discriminator}
\end{align}

To train the generator, we maximize fooling the discriminator.
Additionally, we extract image features and corresponding text features using the pre-trained encoders and compute the DAMSM loss $\mathcal{L}_{\textrm{DAMSM}}(x,t)$ as proposed in \cite{AttnGAN} for an additional learning signal on a per-region level of generated images.
Following \cite{LostGANv2}, we also employ the perceptual loss \cite{ledig2017photo} using extracted VGG \cite{Simonyan2015VeryDC} features and pixel loss for improved image quality.
With our multi-modal region feature matching loss, the generator objective becomes:

\begin{equation}\label{eq:generator}
\begin{aligned}
  \mathcal{L}_{G}=-&\lambda_1\mathbb{E}[D_X(x_\textrm{fake})]-\lambda_2\mathbb{E}[D_R(r_\textrm{fake},t_\textrm{match})]\\
  +&\mathcal{L}_{\textrm{DAMSM}}(x_\textrm{fake},t_\textrm{match})\\
  +&\mathcal{L}_{\textrm{MMRFM}}(r_\textrm{real},r_\textrm{fake},t_\textrm{match})\\
  +&\mathcal{L}_{\textrm{VGG}}(x_\textrm{real},x_\textrm{fake})+\mathcal{L}_{\textrm{pixel}}(x_\textrm{real}-x_\textrm{fake})
\end{aligned}
\end{equation}

Our code is based on the official repositories of \cite{LostGAN,LostGANv2,frolov2021attrlostgan}, and we use the pre-trained BERT model from \cite{wolf-etal-2020-transformers} as our text encoder.
For training, we use the Adam \cite{kingma2014adam} optimizer with $\beta_1=0.0$ and $\beta_2=0.999$.
The learning rates are set to $10e^{-4}$, and weights are fixed as in \cite{LostGAN} to $\lambda_1=0.1, \lambda_2=1.0$.
We train our models using a batch size of 128 on 4 NVIDIA V100-32GB GPUs for 200 epochs which takes roughly 12 days using an image resolution of 128$\times$128.

\section{Experiments}
\label{sec:experiments}

\subsection{Synthetic Images}

\myparagraph{Dataset}
We first create a synthetic dataset similar to \cite{liu2021learning} based on CLEVR \cite{johnson2017clevr} to qualitatively validate our approach.
To that end, we render 50,000 images each depicting 3-8 objects where each object consists of four different attributes including color, shape, size, and material.
Using these attributes, we can create simple text descriptions of localized objects (\eg ``a small red cube'').
To simulate complex image regions with multiple objects, we randomly group nearby objects and additionally use spatial relationship annotations to describe them (\eg ``a small yellow sphere \textit{behind} a small green cylinder'').

\begin{figure}[tb]
  \centering
    \includegraphics[width=\linewidth]{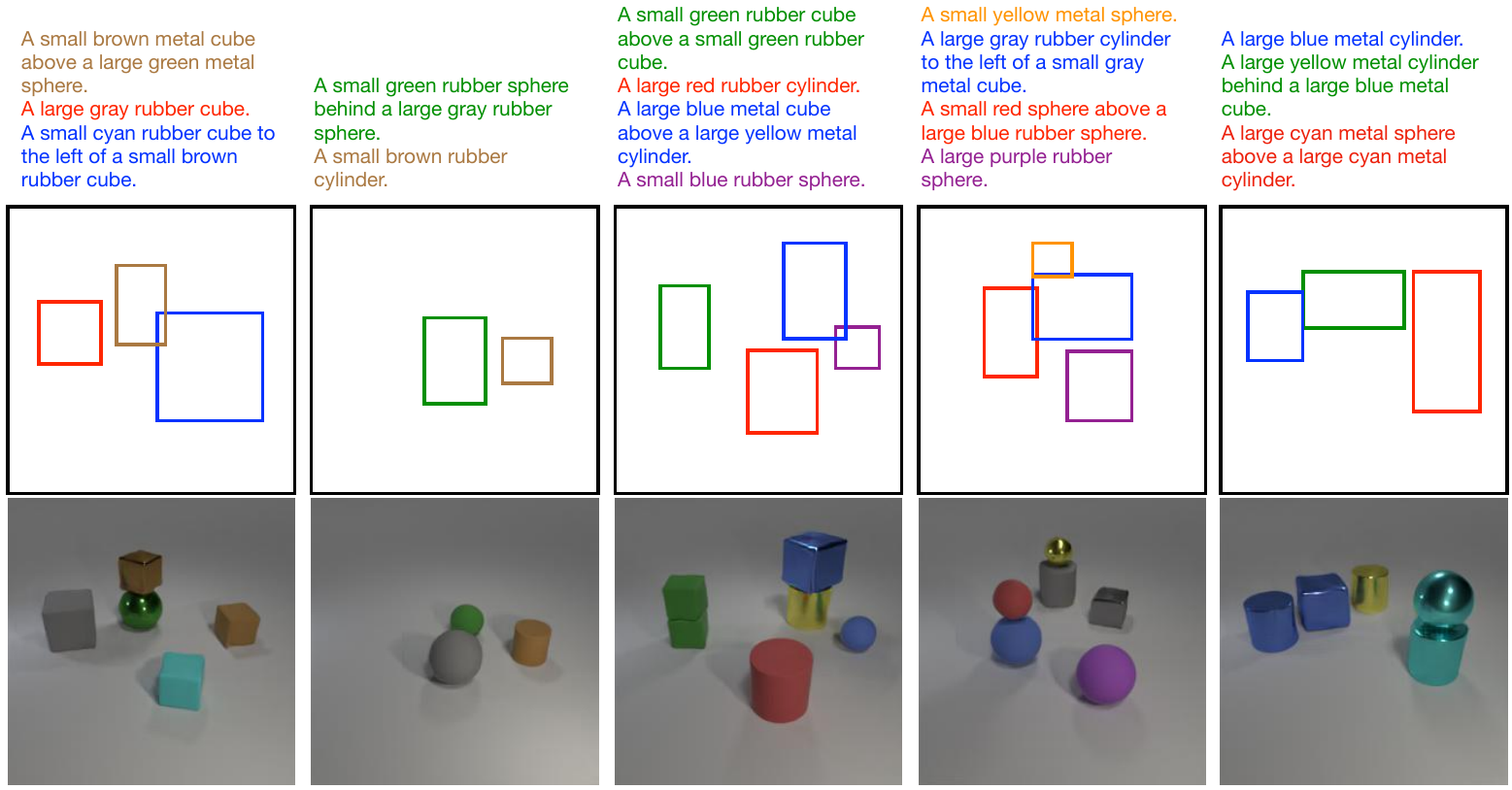}
    \caption{Results on CLEVR.
            Our model can faithfully produce image regions with correct objects, attributes and relationships.}
    \label{fig:clevr}
\end{figure}

\myparagraph{Results}
\cref{fig:clevr} shows input layouts with text descriptions and the corresponding images as generated by our model.
As can be seen, the image quality is high as every object is clearly visible and generated according to the input description.
Our model learned to faithfully produce image regions that describe two objects in a spatial relationship.
In contrast to \cite{liu2021learning}, our model allows to precisely define the location of generated objects and in particular generates only what is specified as input.

\begin{figure}[tb]
  \centering
    \includegraphics[width=\linewidth]{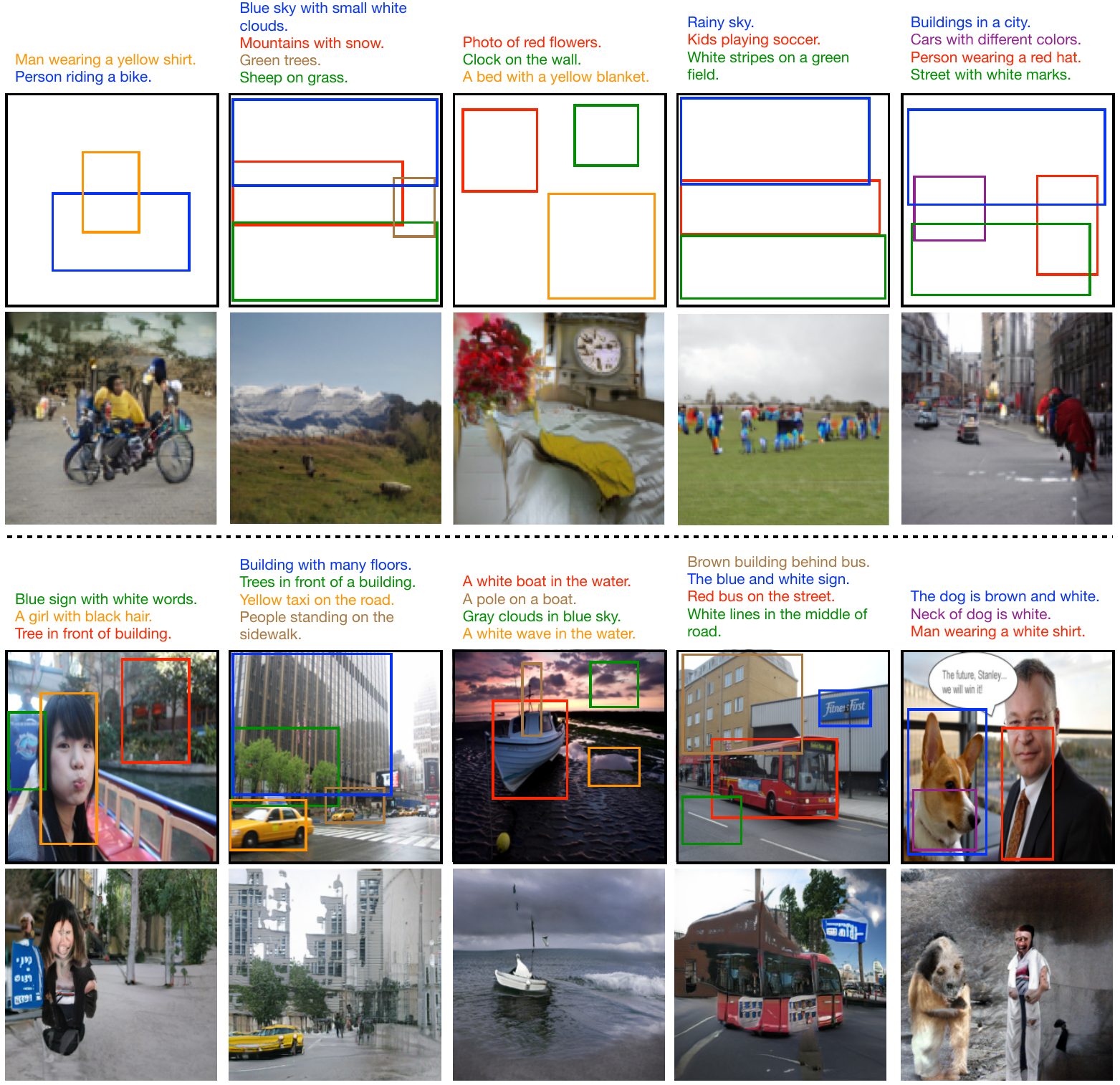}
    \caption{Results on Visual Genome.
            Images generated by manual annotation (top), and from the test set with real images for comparison (bottom).
            Our model can often generate plausible images, correctly respecting the input descriptions.}
    \label{fig:vg}
\end{figure}

\subsection{Real Images}

\myparagraph{Dataset}
We use the Visual Genome (VG) \cite{VisualGenome} dataset, which contains complex images with multiple, interacting objects and region descriptions that describe localized portions of the image.
The region descriptions in the VG dataset are very noisy, repetitive and contain many small regions.
We remove regions smaller than 32$\times$32 pixels and only keep images that contain 3-30 region descriptions with a minimum description length of 5 words.
During training, we select a maximum of 10 regions randomly.
Next, we find and remove similar region descriptions if they are overlapping for more than 30\% and have similar area size.
In summary, we get 85,414 train, 6,206 validation, 6,173 test images, and a total of 1,117,766 region descriptions.

\myparagraph{Evaluation Metrics}
A good DT2I model should produce realistic images and image regions that reflect the semantic meaning of the input captions.
Using several commonly used measures, we extensively evaluate our model in terms of image quality and image-text alignment.

\mysubparagraph{Image Quality}
The Inception Score (IS) \cite{IS} uses a pre-trained image classifier to measure recognizability as well as image diversity of generated images.
We also compute the Fréchet Inception Distance (FID) \cite{FID} which measures the distance between feature distributions of real and generated images. %
Following \cite{OCGAN,frolov2021attrlostgan}, we evaluate the visual fidelity of individual image regions by applying the IS and FID on image crops, denoted as SceneIS \cite{frolov2021attrlostgan} and SceneFID \cite{OCGAN}.

\mysubparagraph{Image-Text Alignment}
To evaluate the semantic matching between input captions and generated images, we adopt the R-precision \cite{AttnGAN} by concatenating all captions per image as a full-scene description.
We report top-5 accuracy given that our task requires distinguishing between multiple text-described regions which is much more challenging when compared to traditional T2I methods.
Furthermore, we use the recently proposed CLIP-Score \cite{hessel2021clipscore} to assess global image-text compatibility using a general-purpose model trained on a large dataset.

\begin{table*}[tb]
\setlength{\tabcolsep}{3pt}
\small
\begin{center}
    \begin{tabular}{l c c c c c c}
    \toprule
    Method 128$\times$128 & IS $\uparrow$ & SceneIS $\uparrow$ & FID $\downarrow$ & SceneFID $\downarrow$ & R-prec. $\uparrow$ & CLIP-S $\uparrow$ \\
    \midrule
    LostGAN \cite{LostGAN} & 11.10 & - & 29.65 & 13.17 & - & - \\
    LostGANv2 \cite{LostGANv2} & 10.71 & - & 29.00 & - & - & - \\
    AttrLostGANv2 \cite{frolov2021attrlostgan} & 10.81 & \bftab 9.46 & 31.57 & \bftab 7.78 & - & - \\
    OC-GAN \cite{OCGAN} & 12.30 & - & 28.26 & 9.63 & - & - \\
    LAMA \cite{li2021image} & - & - & 23.02 & 8.28 & - & - \\ 
    CAL2IM \cite{he2021context} & 12.69 & - & \bftab 21.78 & - & - \\ 
    \midrule
    DTC-GAN (our base) & 8.65 & 8.00 & 39.80 & 9.58 & 55.93 & 63.98 \\
    + triplet & 8.41 & 7.89 & 44.39 & 11.86 & 59.97 & 66.56 \\
    + MMRFM  & 8.43 & 8.30 & 47.08 & 13.63 & 58.50 & 66.19 \\
    + DAMSM (our final) & \bftab 13.80 & \bftab 9.46 & 27.67 & \underline{7.85} & \bftab 89.07 & \bftab 68.20  \\
    \midrule
    \midrule
    Method 256$\times$256 & IS $\uparrow$ & SceneIS $\uparrow$ & FID $\downarrow$ & SceneFID $\downarrow$ & R-prec. $\uparrow$ & CLIP-S $\uparrow$ \\
    \midrule
    LostGANv2 \cite{LostGANv2} & 14.10 & - & 47.62 & - & - & - \\
    AttrLostGANv2 \cite{frolov2021attrlostgan} & 14.25 & 11.96 & 35.73 & 14.76 & - & - \\
    OC-GAN \cite{OCGAN} & 14.70 & - & 40.85 & - & - & - \\
    LAMA \cite{li2021image} & - & - & \bftab 31.63 & 13.66 & - & - \\ 
    \midrule
    DTC-GAN (our final) & \bftab 15.96 & \bftab 12.03 & \underline{35.29} & \bftab 9.32 & \bftab 77.68 & \bftab 67.98  \\
    \bottomrule
    \end{tabular}
    \end{center}
    \caption{Quantitative results on Visual Genome.
    Our model benefits from combining all discussed losses and outperforms previous layout-to-image methods on several metrics while allowing free-form region descriptions as input.
    Both \cite{li2021image} and \cite{he2021context} are recent improvements which could also be applied to our backbone generator.
    }
\label{tab:vg}
\end{table*}

\myparagraph{Results}
\cref{fig:vg} shows images generated by our model.
Our method provides an unprecedented interface for the user to communicate what image should be generated.
Despite the very challenging task it is  able to produce images which match the input region descriptions with objects at the correct location.
Furthermore, the model respects color attributes and compositions such as ``man wearing a yellow shirt'' and ``mountains with snow''.
Next, we evaluate and compare our model with several recent methods, see \cref{tab:vg}.
A naive replacement of labels with text embeddings (our base) leads to low scores across all metrics.
Although adding the MMRFM loss results in slightly lower scores, we observed faster and stable convergence during training and better performance when combining all losses.
Combining two techniques from the T2I literature with our proposed MMRFM loss outperforms previous methods on several metrics while allowing free-form region descriptions as input.

\myparagraph{Limitations}
Generating images of complex scenes is a very challenging task.
As all current approaches, our model still struggles to produce highly realistic images.
From a training perspective, our model relies on dense captions, which are time-consuming to collect.
The used dataset mostly contains rather short captions, and it would be interesting to expand towards longer descriptions with more details.
During testing, we observed that details such as positional and numerical information is often ignored during image generation, which is also currently a problem in the wider T2I literature \cite{frolov2021adversarial}.
Improved mask prediction \cite{OCGAN,li2021image} and context awareness \cite{he2021context}, both of which are recent improvements of the generator, may as well improve the image quality for our application.
While our method produces promising results, it is merely the first step in this direction and further research efforts are required to improve the image quality to enable practical applications.

\section{Conclusion}
\label{sec:conclusion}
In this paper, we introduced dense text-to-image (DT2I) synthesis as a new task and proposed DTC-GAN, a novel method that generates images from multiple free-form region descriptions.
To address this challenging task, our method successfully combines recent layout-to-image and text-to-image techniques.
Furthermore, we proposed a multi-modal region feature matching loss to improve semantic image-text matching between real and generated image regions and stabilize training.
From a user's perspective, our model is the first of a new kind of image generation method which are intuitive, flexible, and not restricted by a fixed set of labels.
In terms of future work, enhancing the quality of generated images is required for practical applications and can for example be achieved by scaling up the networks and using contrastive losses.

\myparagraph{Acknowledgments} 
This work was supported by the BMBF projects
ExplAINN (Grant 01IS19074),
XAINES (Grant 01IW20005),
the NVIDIA AI Lab (NVAIL)
and the TU Kaiserslautern PhD program.

\bibliographystyle{splncs04}
\bibliography{egbib}

\end{document}